\documentclass[10pt,twocolumn,letterpaper]{article}

\usepackage[algorithms]{wacv}      
\usepackage{wacv}              

\usepackage{graphicx}
\usepackage{amsmath}
\usepackage{amssymb}
\usepackage{booktabs}

\usepackage[pagebackref,breaklinks,colorlinks]{hyperref}

\usepackage[capitalize]{cleveref}
\crefname{section}{Sec.}{Secs.}
\Crefname{section}{Section}{Sections}
\Crefname{table}{Table}{Tables}
\crefname{table}{Tab.}{Tabs.}

\def\wacvPaperID{1401} 
\def\confName{WACV}
\def\confYear{2025}

\begin{document}

\title{Event-guided Low-light Video Semantic Segmentation}

\author{Zhen Yao\\
Lehigh University\\
{\tt\small zhy321@lehigh.edu}
\and
Mooi Choo Chuah\\
Lehigh University\\
{\tt\small mcc7@lehigh.edu}
}
\maketitle

\begin{abstract}
   Recent video semantic segmentation (VSS) methods have demonstrated promising results in well-lit environments. However, their performance significantly drops in low-light scenarios due to limited visibility and reduced contextual details. In addition, unfavorable low-light conditions make it harder to incorporate temporal consistency across video frames and thus, lead to video flickering effects. Compared with conventional cameras, event cameras can capture motion dynamics, filter out temporal-redundant information, and are robust to lighting conditions. To this end, we propose EVSNet, a lightweight framework that leverages event modality to guide the learning of a unified illumination-invariant representation. Specifically, we leverage a Motion Extraction Module to extract short-term and long-term temporal motions from event modality and a Motion Fusion Module to integrate image features and motion features adaptively. Furthermore, we use a Temporal Decoder to exploit video contexts and generate segmentation predictions. Such designs in EVSNet result in a lightweight architecture while achieving SOTA performance. Experimental results on 3 large-scale datasets demonstrate our proposed EVSNet outperforms SOTA methods with up to $11\times$ higher parameter efficiency.
\end{abstract}

\section{Introduction}\label{sec:intro}
Video semantic segmentation, a problem of assigning a category label to each pixel in the video frames, has become a hot research topic in recent years. It plays a fundamental role in a wide range of multimedia and computer vision applications including video parsing \cite{li2024endora,li2024feature}, video processing \cite{bai2024anything, ying2021srnet, li2022video}, and autonomous driving \cite{yang2024monogae, zhao2024roadbev, zhao2024road, huang2024toward}. \par

\begin{figure}[!ht]
  \includegraphics[width=0.48\textwidth]{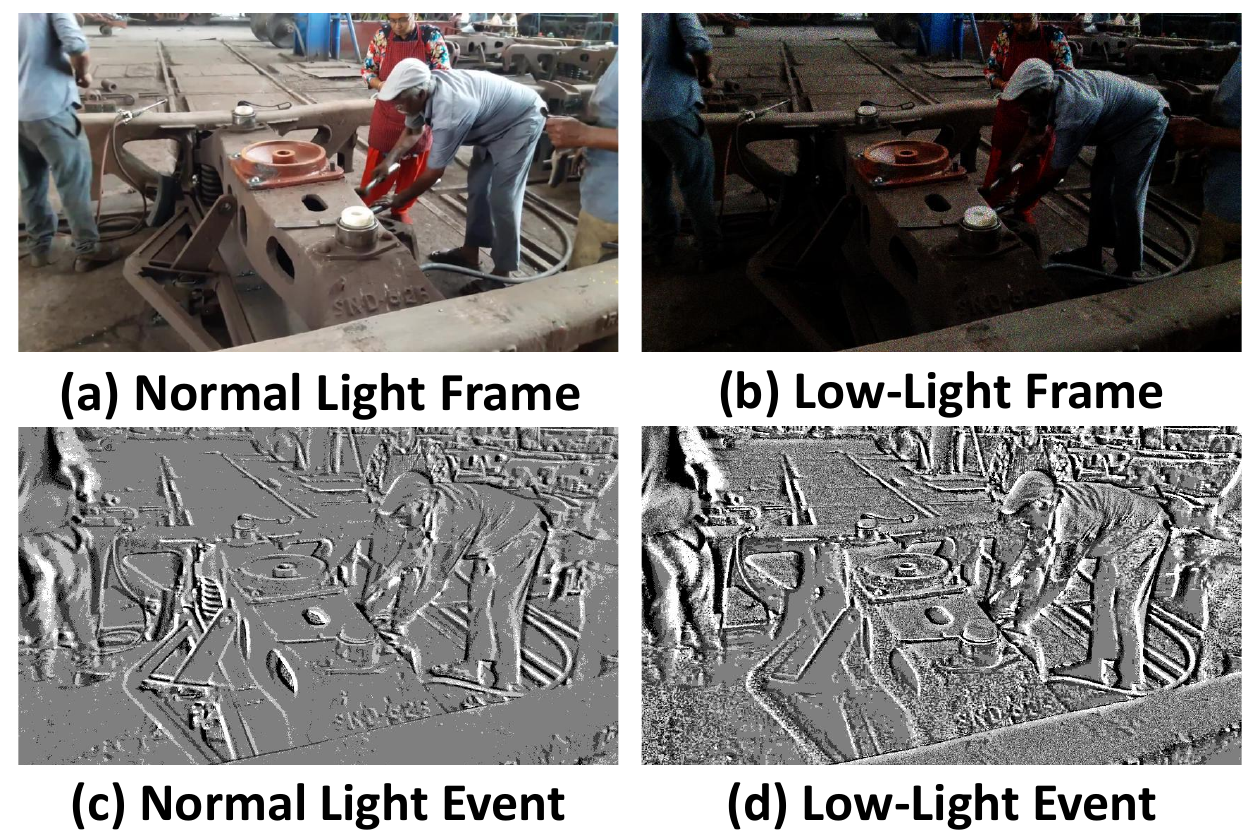}
  \caption{Comparison among (a) normal light frame, (b) low-light frame, (c) events generated from normal light frames, and (d) events generated from low-light frames. Events demonstrate robustness against lighting changes and effectively capture temporal motion features in low-light environments.}
  \label{fig:teaser}
\vspace{-0.4cm}

\end{figure}

While video semantic segmentation of normal light scenes has made tremendous achievements \cite{sun2022coarse, li2022video, miao2021vspw, wang2021temporal, zhao2017pyramid}, low-light scenarios are still challenging due to limited visibility and degraded image quality. In low-light conditions, conventional frame-based cameras have difficulty capturing a wide range of brightness levels, resulting in low contrast and loss of textures. The reduced contrast hinders accurate discrimination of object boundaries, ultimately diminishing clarity and fidelity of the captured images. In addition to limited visibility, low-light semantic segmentation is also challenging because of luminance noise. Severe noises caused by constrained photon counts and imperfections in photodetectors manifest as random bright or dark pixels scattered throughout the whole image. 
Such noises often lead to inaccurate segmentation predictions.
\par

To resolve the above issues, researchers have explored event cameras as an alternate sensing modality. Event sensors asynchronously measure sparse data streams at high temporal resolution (10µs vs 3ms), higher dynamic range (140dB vs 60dB), and significantly lower energy (10mW vs 3W) compared to conventional cameras. In recent years, it has been increasingly utilized in the computer vision \cite{gehrig2023recurrent,li2021graph,chen2022efficient,huang2024multi,sun2023hmaac,jia2022molecular} and robotics \cite{wang2022ev,vitale2021event} community. Researchers have explored event modality in many tasks such as 3D reconstruction \cite{zhou2018semi}, pose estimation \cite{chen2022efficient}, and simultaneous localization and mapping (SLAM) \cite{guo2024cmax}. \par

\begin{figure}[!t]
\centering
\includegraphics[width=0.45\textwidth]{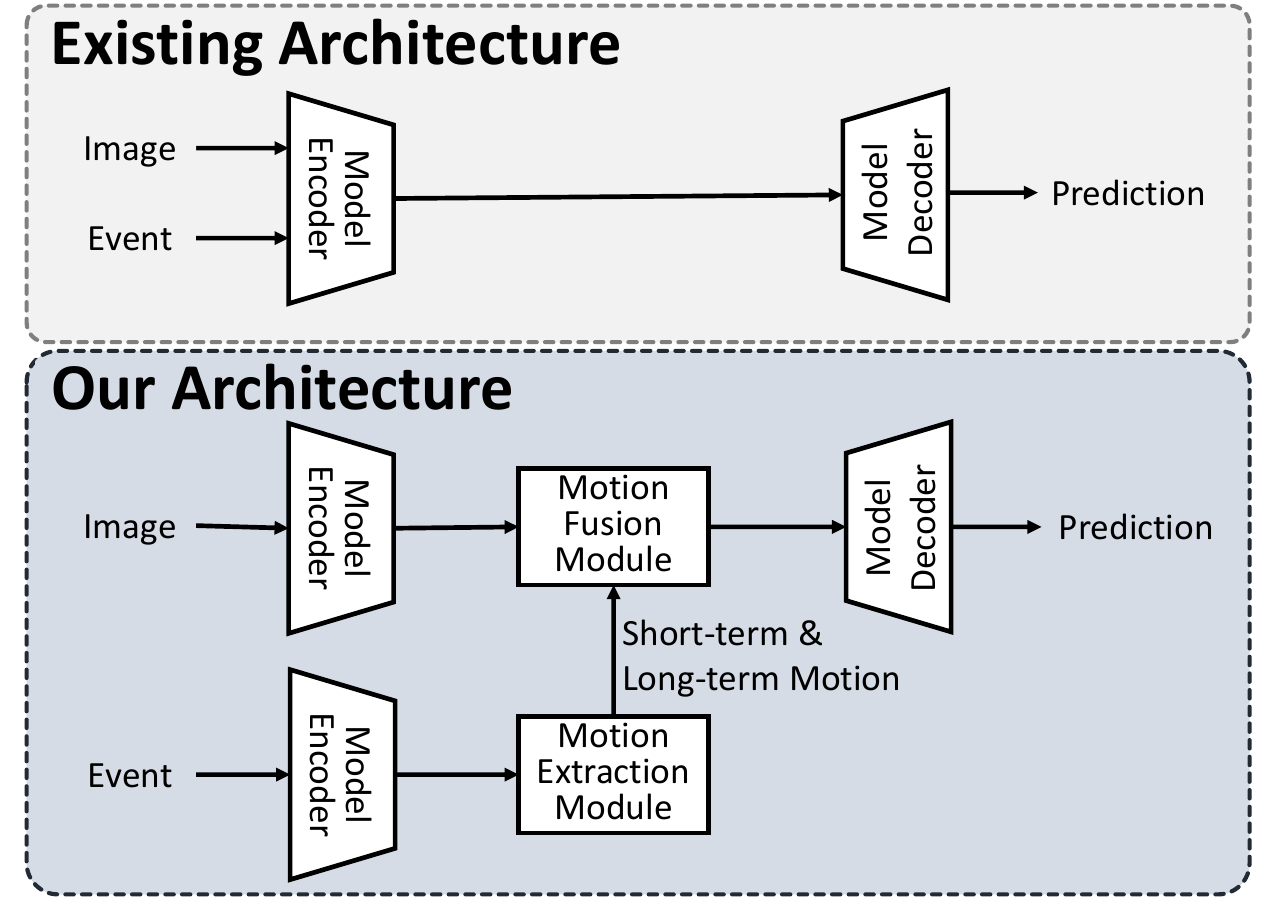}
\caption{Different from existing approach which fuse event features in the encoder, our model utilizes event features to learn short-term and long-term motions and leverage these motion features to gain video temporal contexts.}
\label{diff}
\vspace{-0.4cm}

\end{figure}

Instead of capturing an image at a fixed interval, the event cameras, such as the Dynamic Vision Sensor (DVS), only record a single event based on the brightness changes at each pixel. This makes it suitable for edge-case scenarios and video-related tasks. For low-light video semantic segmentation task, event modality offers two advantages: (1) it facilitates the learning of temporal consistency; (2) it provides an alternative perspective through intensity changes. Figure. \ref{fig:teaser} explains the potential of event modality in low-light conditions. Although bringing a new paradigm shift, event modality presents two challenges: firstly, it only captures pixels in motion, leading to sparse information; secondly, it demonstrates distinct attributes compared with visual image modality, highlighting the significance of effectively modeling event features. \par

To address the above challenges, we propose a lightweight framework, namely EVSNet, that exploits both image and event modalities for low-light video semantic segmentation. Figure. \ref{diff} illustrates the difference between existing works and our approach. The architecture of EVSNet is shown in Figure. \ref{arch}. It consists of three parts: an Image Encoder, a Motion Extraction Module, a Motion Fusion Module, and a Temporal Decoder. Specifically, we first adopt a lightweight backbone as the Image Encoder to extract image features. Inspired by Atkinson-Shiffrin memory model \cite{atkinson1968human} which hypothesizes that the human memory consists of short memory and long memory, we then propose a Motion Extraction Module to extract short-term and long-term motion features and acquire the video contexts to guide semantic understanding through such motions. We apply Event Encoder to extract event features (can be seen as short-term motions between 2 consecutive frames) and leverage the Temporal Convolutional Block to learn long-term motions. Furthermore, we devise a Motion Fusion Module to integrate image and motion features adaptively. It contains a Channel Attention layer and a Spatial Attention Layer to blend cross-channel and spatial information. By leveraging both modalities, EVSNet extracts richer and complementary information, leading to more accurate segmentation compared to the single modality. Finally, we use a Temporal Decoder to exploit video contexts and generate final predictions. Extensive evaluations using three large-scale low-light datasets show EVSNet results in better semantic segmentation results. \par

In summary, our contributions to this paper include:
\begin{itemize}
\item We propose EVSNet, a lightweight framework that exploits image and event modality. To the best of our knowledge, we are the first to introduce event modality to video semantic segmentation task. Event information focuses on the motion changes and thus can be used to learn better short and long-term temporal consistent representations. 
\item We propose a Motion Extraction Module (MEM) and a Motion Fusion Module (MFM) for learning temporal motion and adaptively learning the spatial and channel-wise relationship between image and motion features. Unlike existing extraction and fusion modules, our design alleviates misalignment while lowering computational cost.
\item We conduct experiments to evaluate our EVSNet model using three large-scale low-light video semantics segmentation datasets and demonstrate its effectiveness using standard segmentation metrics. Compared to SOTA models with similar parameter efficiency and inference cost, our EVSNet achieves superior performance on these 3 datasets.
\end{itemize}
\section{Related Works}\label{sec:related}
\subsection{Event-Based Semantic Segmentation}
Event cameras have the potential for semantic segmentation and there have been several efforts exploring event modality. Some researchers use knowledge distillation \cite{qiu2023sats} to solve domain shift problems. ESS \cite{sun2022ess} proposed to leverage unsupervised domain adaptation by aligning event features with image features. EvDistill \cite{wang2021evdistill} designed a student network on unlabeled event data to distill knowledge from a teacher network trained with labeled data. Some researchers use other ways to encode event features. Ev-SegNet \cite{alonso2019ev} built an Xception-based CNN to train on event data. HALSIE \cite{biswas2022halsie} proposed to use hybrid spiking neural networks \cite{hsieh2024bio} and convolution neural networks to extract spatiotemporal features to combine events and frames. \par

While researchers started to exploit event modality for semantic segmentation, most works \cite{chen2019appearance,li2024event} didn't explore how to effectively fuse the multimodality and alleviate misalignment. Unlike existing approaches, our design learns longer term motion features from event data, and fuse such features with image features at a later stage. Such a design yields better segmentation results.

\subsection{Low-light Semantic Segmentation}
Low-light semantic segmentation has been popular in recent years. Due to the absence of a real-world low-light segmentation dataset in early stages, some previous methods \cite{wang2024unsupervised, xia2023cmda} utilized domain adaptation to transfer knowledge from the normal light domain to the low-light domain. DANNet \cite{wu2021one} designed a domain adaptation network utilizing an annotated normal light dataset as the source domain and an unlabeled dataset that contains coarsely matched image pairs (the target normal light and low-light domains). It used multi-target adaptation and re-weighting strategy to enhance the accuracy. LISU \cite{zhang2022lisu} devised a cascade framework to enhance segmentation predictions of low-light scenarios by jointly learning semantic segmentation and reflectance restoration. \cite{fan2020integrating} utilized semantic segmentation as guidance to help the Retinex-based model learn low-light image enhancement based on structural and semantic prior. It reduces noise and color distortion and improves visual quality in low-light environments. \par

\subsection{Video Semantic Segmentation}
Video semantic segmentation, compared to image segmentation, exploits temporal information in consecutive frames leveraging motion cues and temporal context. Some researchers \cite{xu2018dynamic,nilsson2018semantic,jain2019accel,lee2021gsvnet,gadde2017semantic} utilized optical flow to warp features from frames and then aggregate wrapped features. EVS \cite{paul2020efficient} proposed a novel Refiner to Warp semantic information and IAM focusing on regions where the optical flow is unreliable. Accel \cite{jain2019accel} proposed to use a reference branch for extracting fine-grained features from keyframes and warping features using optical flow, and an update branch for performing a temporal update on the current frame. DEVA \cite{cheng2023tracking} developed bi-directional propagation fusing of segmentation hypotheses and current segmentation results to predict coherent labels. Further works focus on how to model temporal consistency between frames \cite{sun2022coarse,lao2023simultaneously,paul2021local,sun2022mining,hu2023efficient}. CFFM \cite{sun2022coarse} proposed coarse-to-fine feature assembling and cross-frame feature mining. The former extracted fine-grain and coarse-grain features while the latter mined temporal relations based on focal features. Video K-Net \cite{li2022video} proposed a unified framework for multiple video segmentation tasks but requires separate training for each task. \par

Despite the promising results, existing approaches do not learn temporal motion features well. NetWarp \cite{gadde2017semantic} wraps previous frame to learn temporal motions, but it fails to consider long-term motion contexts. MRCFA \cite{sun2022mining} models cross-frame temporal relations, but it fails to model relation between short-term and long-term temporal motions. In addition, SOTA methods \cite{hu2020temporally,li2021video,paul2021local,qiao2024multi} learn temporal motions from image modality. It is suboptimal because of the complexity of capturing and interpreting dynamic changes from static images over time. Requiring the model to effectively differentiate between spatial and temporal information in low-light conditions is extremely challenging due to both low image contrast and ambiguity of object boundaries. This paper highlights a new direction of mining both short and long-term motions from the event modality and is compatible with most of the encoder-decoder architecture in Video Semantic Segmentation task. \par
\begin{figure*}[!t]
\centering
\includegraphics[width=1.00\textwidth]{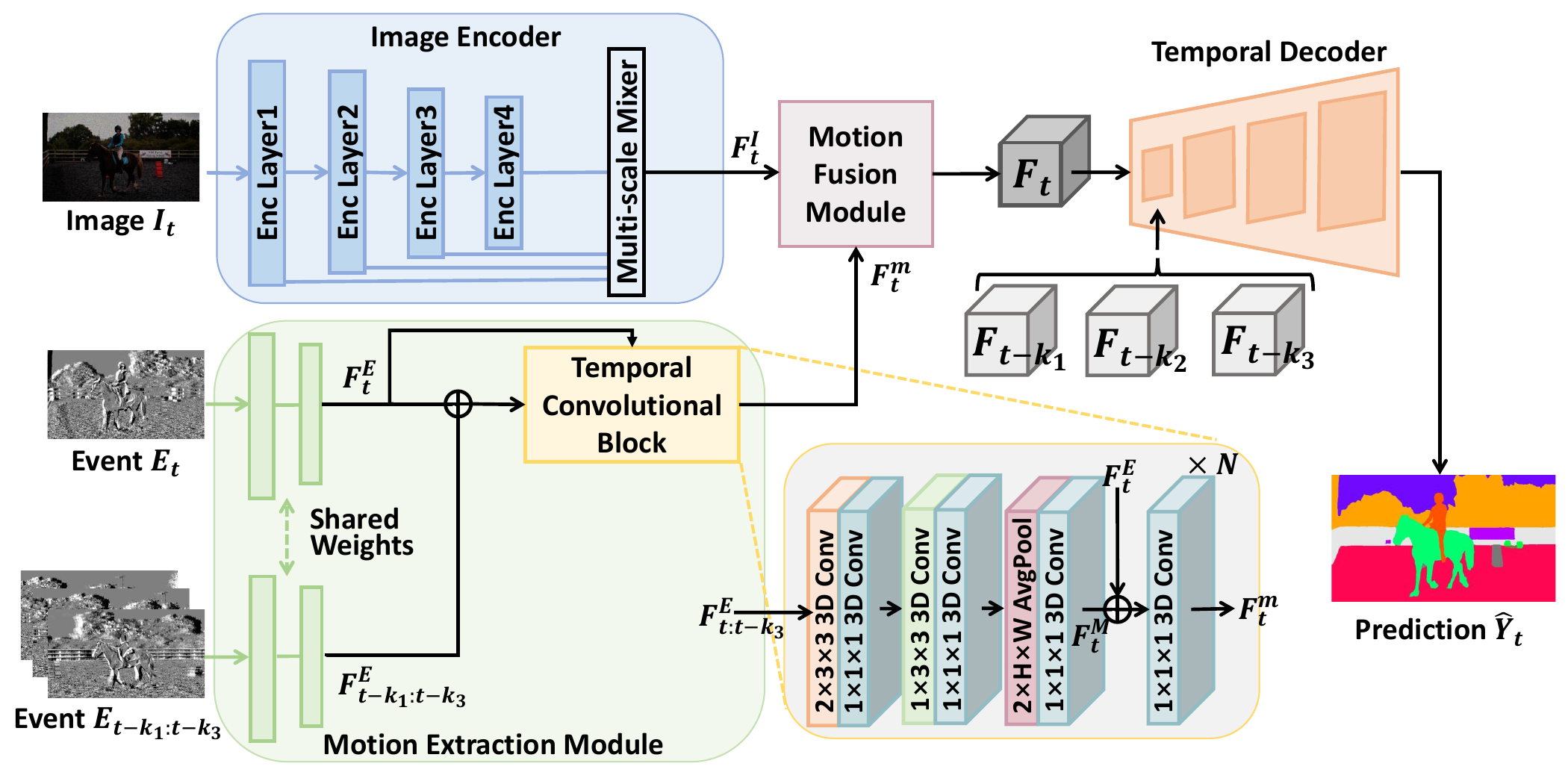}
\caption{Overview of EVSNet: Given a set of input RGB images $\boldsymbol{I_{t:t-k_3}}$, the Image Encoder extracts image features at each scale and multi-scale features are aggregated in the Multi-scale Mixer. The Motion Extraction (MEM) Module is applied to extract short-term and long-term motions from event features generated by event simulator. The Motion Fusion (MFM) Module further aggregates both features adaptively. Finally, the Temporal Decoder combines fused features from all frames to learn the temporal consistency and predicts the final segmentation results.}
\label{arch}
\vspace{-0.2cm}

\end{figure*}

\section{Methodology}\label{sec:method}
\subsection{Motivation}
In low-light scenarios, where visibility is severely compromised, conventional cameras often fail to provide adequate information for robust segmentation due to the low contrast between objects. Furthermore, noise and blur introduced by RGB cameras in low-light environments cause image quality degradation. Video semantic segmentation models designed for normal light scenarios thus cannot clearly capture low-level and high-level features in each video frame and have difficulty acquiring scene
understanding capabilities \cite{chen2023instance}. Adding extra denoising/restoration steps or modules brings excessive computational cost and exacerbates the overall latency of the pipeline. \par

The event modality, characterized by its ability to capture dynamic changes (such as motion and sudden illumination alterations) in the scene, offers valuable structural and motional information that is not captured by conventional cameras. It naturally responds to motion and is especially suitable for video-related tasks. By integrating the event modality alongside image modality, the model exploits complementary sources of richer semantic information, understands more comprehensively about the environment, and ultimately enhances segmentation accuracy and robustness. Hence, the model can better adapt to challenging low-light conditions and offer a promising avenue for improving performance and applicability in real-world scenarios. \par

\subsection{Problem Formulation}
\vspace{-0.1in}
Assuming there is an input video clip containing $\boldsymbol{l+1}$ video frames $\boldsymbol{\{I_{t}, I_{t-k_{1}}, ..., I_{t-k_{l}}\}} \in \mathbb{R}^{H \times W \times 3}$ with corresponding per-frame ground-truth segmentation label $\boldsymbol{\{S_{t}, S_{t-k_{1}}, ..., S_{t-k_{l}}\}} \in \mathbb{R}^{H \times W \times 1}$. Note that $\boldsymbol{I_{t}}$ is referred as current frame and $\boldsymbol{\{I_{t-k_{1}}, ..., I_{t-k_{l}}\}}$ are $l$ previous frames which are
$\boldsymbol{\{k_{1}, ..., k_{l}\}}$ frames away from $\boldsymbol{I_{t}}$. They are defined as reference frames. Our objective is to make pixel-wise segmentation on $\boldsymbol{I_{t}}$. \par

Event cameras capture the changes in intensities for each pixel and output a stream of events. One event $\boldsymbol{e_i}$ is represented as the 4-tuple: $e_i = \left[x_i, y_i, p_i, t_i\right]$ where $x_i, y_i$ indicates the spatial coordinates of the pixel where the brightness changes at timestamp $t_i$ and polarity $p_i \in \{0, 1\}$ denotes either increasing or decreasing of the local brightness. Event simulators \cite{hu2021v2e} usually transform the asynchronous event flows into synchronous event image by stacking the events in a fixed time interval $\Delta t$. Similar to \cite{nguyen2017real}, we encode the event information as a one-channel grey-scale image to facilitate the learning of event modality. Event images of the given video clip are referred as $\boldsymbol{\{E_{t}, E_{t-k_{1}}, ..., E_{t-k_{l}}\}} \in \mathbb{R}^{H \times W \times 1}$ for each corresponding frame. \par

\subsection{Method Overview}
The proposed framework of EVSNet, as illustrated in Figure. \ref{arch}, consists of an Image Encoder for feature extraction from image modality, a Motion Extraction Module (MEM) for extracting temporal motions from event modality, a Motion Fusion Module (MFM) for feature integration, and a Temporal Decoder for modeling temporal relations between frames and generating pixel-wise predictions. \par

\textbf{Image Encoder.} Given a current video frame $\boldsymbol{I_t} \in \mathbb{R}^{H \times W \times 3}$, the Image Encoder utilizes a lightweight pretrained backbone to extract multi-scale features and reduce the computational overhead. Specifically, we adopt two efficient architectures: Afformer (Base and Tiny) \cite{dong2023head} and MiT (B0 and B1) \cite{xie2021segformer}. In addition, the low-pass filter design in the backbone can help suppress noise generated in low-light conditions \cite{chen2023instance}. Note that the Image Encoder extracts feature maps from the current frame $\boldsymbol{I_t} \in \mathbb{R}^{H \times W \times 3}$ and reference frames $\boldsymbol{\{I_{t-k_{1}}, ..., I_{t-k_{l}}\}} \in \mathbb{R}^{H \times W \times 3}$, respectively. Finally, we leverage the MLP decoder in SegFormer \cite{xie2021segformer} as the Multi-scale Mixer to aggregate multi-scale features $\boldsymbol{F^I_{t}} \in \mathbb{R}^{H/4 \times W/4 \times C}$ where local features from earlier layers can be combined well with global features from the latter layers. \par

\textbf{Motion Extraction Module.} Similar to the Image Encoder, the Motion Extraction Module (MEM) uses a pretrained backbone on the event images $\boldsymbol{\{E_{t}, ..., E_{t-k_{l}}\}} \in \mathbb{R}^{H \times W \times 1}$ to generate event features $\boldsymbol{\{F^E_{t}, ..., F^E_{t-k_{l}}\}} \in \mathbb{R}^{H \times W \times C}$. Subsequently, we pass event features through the Temporal Convolutional block to learn the long-term temporal relations. The Temporal Convolutional block extracts motion feature maps from the current and past $l$ event frames to generate motion features $\boldsymbol{F_t^m}$. For instance, for the current frame $t$, it outputs long-term motion features $\boldsymbol{F^m_{t}} \in \mathbb{R}^{H \times W \times C}$ based on $\boldsymbol{\{E_{t}, ..., E_{t-k_{l}}\}}$. More details of the Motion Extraction Module are described in Section \ref{sec:mem}. \par

\textbf{Motion Fusion Module.} The output of the Image Encoder and Motion Extraction Module are passed through a Motion Fusion Module (MFM). Specifically, we apply a Channel Attention Layer to learn inter-channel dependencies of both features and then a Spatial Attention Layer to learn spatial structural details. It then feeds the updated features to the decoder. The MFM is lightweight and efficient while yielding a powerful representation, incorporating the short-term and long-term temporal contexts. More details of the Motion Fusion Module are described in Section \ref{sec:fusion}. \par

\textbf{Temporal Decoder.} The Temporal Decoder takes the fused features from the image and motion features as input. The Temporal Decoder consists of several focal blocks from Focal Transformer \cite{yang2021focal}. It is applied to learn the temporal consistency of consecutive frames and further predicts the segmentation results $\boldsymbol{\hat{Y}_{t}} \in \mathbb{R}^{H \times W \times 1}$. More details of Temporal Decoder selections are elaborated in Section \ref{sec:ablation}. \par

\begin{figure}[!t]
\centering
\includegraphics[width=1.05\linewidth]{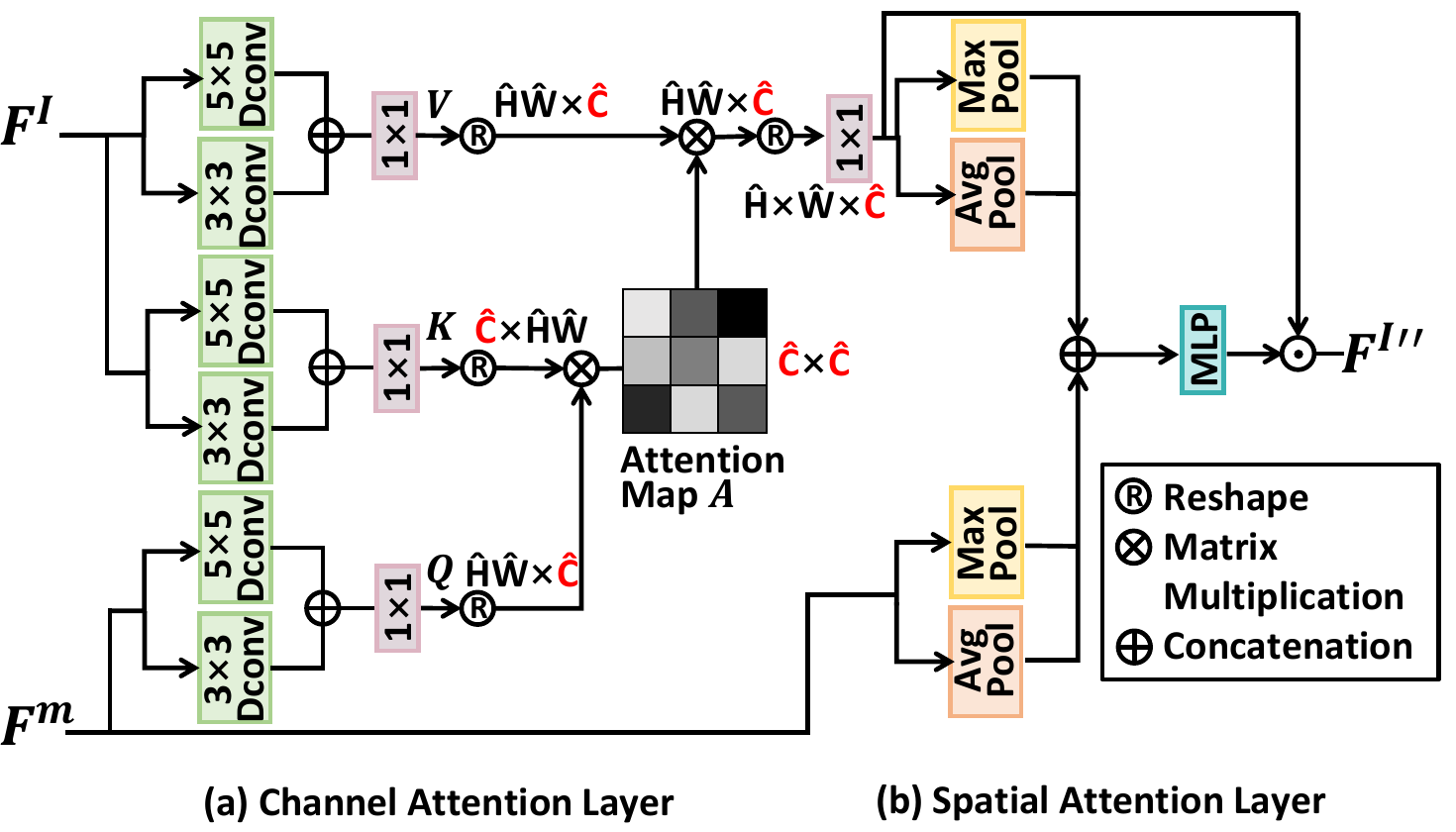}
\caption{Illustration of Motion Fusion Module (MFM): From left to
right, the components are Channel Attention Layer and Spatial Attention Layer.}
\label{fig:mfm}
\vspace{-0.4cm}
\end{figure}

\subsection{Motion Extraction Module}\label{sec:mem}
The image features contain more semantic and category-aware information while the event features contain more structural and category-agnostic information \cite{liao2024mobile}. There is a significant domain gap in the two modalities and how to alleviate the domain shift problem is important \cite{li2022domain,li2021consistent,wang2024mixsegnet,wang2023dual,wei2022lidar,zhu2023topic,wang2024landa,wang2022metateacher}. Some prior works directly fuse features from multiple scales or streams \cite{biswas2022halsie}, and other researchers introduce atrous spatial pyramid pooling or pyramid feature fusion modules to encode spatial dependencies \cite{yao2024cracknex}. However, pixels corresponding to the same scene points in the event and RGB images are recorded in different temporal resolutions, leading to severe misalignment, and this misalignment cannot be easily solved by multimodality fusion. \par

Based on the analysis of existing fusion modules above, we summarize the following design principles for our model: (1) The event modality is mainly used to extract temporal motions. Given that the event records brightness changes between two consecutive frames, it can be seen as short-term temporal motions. Inspired by Atkinson-Shiffrin memory model \cite{atkinson1968human}, we also leverage event modality to learn long-term temporal motions. (2) Directly integrating event features in RGB Image Encoder introduces misalignment and may affect the semantic understanding. Instead, we extract motion features and utilize such information to gain video contexts. \par

The Motion Extraction Module consists of 2 parts: Event Encoder and Temporal Convolutional block. Given the input event frames $\boldsymbol{\{E_{t}, ..., E_{t-k_{l}}\}} \in \mathbb{R}^{H \times W \times 1}$, the Event Encoder extracts multi-scale high-level features $\boldsymbol{\{F^E_{t}, ..., F^E_{t-k_{l}}\}} \in \mathbb{R}^{H \times W \times C}$. The event features $\boldsymbol{F^E}$ are then fed to the Temporal Convolutional block. The block consists of a $2\times3\times3$ 3D Convolution layer, a $1\times3\times3$ 3D Convolution layer, a $2\times H\times W$ 3D average pooling layer layer, and skip connections. Each operation above is preceded by a feature compression layer, which is a $1\times1\times1$ 3D Convolutional layer. For a specific frame $t$, the final output motion features $\boldsymbol{F^m_t}$ are the fused features after concatenation of short-term event features $\boldsymbol{F^E_t}$ and long-term motion features $\boldsymbol{F^M_t}$ as follows:
\begin{equation} \label{eq:longmotion}
\boldsymbol{F^M_t} = \mathrm{ReLU}(\tau(AvgPool(\tau(\boldsymbol{f_2}(\tau(\boldsymbol{f_1}(\boldsymbol{F^E_t})))) + \boldsymbol{F^E_t})))
\end{equation}
\begin{equation} \label{eq:motion}
\boldsymbol{F^m_t} = \tau(\boldsymbol{F^E_t} \oplus \boldsymbol{F^M_t})
\end{equation}
where $\boldsymbol{f_1}$ and $\boldsymbol{f_2}$ represents $2\times3\times3$ and a $1\times3\times3$ 3D Convolution layer, $\tau$ represents $1\times1\times1$ 3D Convolution
layer and $\boldsymbol{F^m_t} \in \mathbb{R}^{H \times W \times 2C}$ represents the concatenated motion features. \par

\subsection{Motion Fusion Module}\label{sec:fusion}
Based on the two principles in Section \ref{sec:mem}, we propose a lightweight Motion Fusion Module (MFM) to adaptively aggregate image and event features from spatial and channel aspects and improve cross-modal generalization. Specifically, the MFM consists of one Channel Attention layer and one Spatial Attention layer. \par

\textbf{Channel Attention Layer.} The Channel Attention layer is based on the channel-wise attention mechanism \cite{zamir2022restormer} with two key modifications in Figure. \ref{fig:mfm}. (1) Instead of using the image to query and focusing solely on each pixel to learn which channel is more important, we use the event image to query and compute cross-covariance across feature channels. It updates image features with the guide from the event information and focuses on more important channels based on the event modality's motional and structural perspective. (2) We add an additional $5 \times 5$ depth-wise convolution layer in parallel and remove the Feed Forward Networks (FFN) in the original attention blocks. Replacing FFN with the extra depth-wise convolution layer can greatly reduce the computational cost and potential feature misalignment between the two modalities. \par

Given the image $\boldsymbol{F^I}$ and motion feature maps $\boldsymbol{F^m}$, the Channel Attention Layer first uses two parallel depth-wise convolution layers ($3 \times 3$ and $5 \times 5$) and one $1 \times 1$ convolution layer for both inputs. Our design has two advantages: (1) it implicitly models contextual relationships between surrounding pixels; (2) it needs fewer computations than the standard convolutional layer. Afterward, it generates a query vector from the motion features and key/value vectors from image features:
\begin{equation} \label{eq:q}
\boldsymbol{Q} = \boldsymbol{W_Q}(\boldsymbol{f_3}(\boldsymbol{F^m}) \oplus \boldsymbol{f_4}(\boldsymbol{F^m}))
\end{equation}
\begin{equation} \label{eq:k}
\boldsymbol{K} = \boldsymbol{W_K}(\boldsymbol{f_3}(\boldsymbol{F^I}) \oplus \boldsymbol{f_4}(\boldsymbol{F^I}))
\end{equation}
\begin{equation} \label{eq:v}
\boldsymbol{V} = \boldsymbol{W_V}(\boldsymbol{f_3}(\boldsymbol{F^I}) \oplus \boldsymbol{f_4}(\boldsymbol{F^I}))
\end{equation}
where $\oplus$ denotes concatenation and $\boldsymbol{f_3}$ and $\boldsymbol{f_4}$ are $3 \times 3$ and $5 \times 5$ depth-wise convolution layer, respectively. The spatial resolution of input $\boldsymbol{F^I}$ and $\boldsymbol{F^m}$ is $\mathbb{R}^{H \times W \times \hat{C}}$. \par

We then reshape query and key projections such that their dot-product can be multiplied with the reshaped value projections. The updated image features $\boldsymbol{F_I'}$ after the Channel Attention Layer are:
\begin{equation} \label{eq:channelattn}
\boldsymbol{F^{I\prime}} = \boldsymbol{V} \cdot \text{softmax}((\boldsymbol{K} \cdot \boldsymbol{Q}) / \alpha )
\end{equation}
where $\boldsymbol{Q} \in \mathbb{R}^{HW \times \hat{C}}$, $\boldsymbol{K} \in \mathbb{R}^{\hat{C} \times HW}$ and $\boldsymbol{V} \in \mathbb{R}^{HW\times \hat{C}}$ are reshaped tensors originally from the size $\mathbb{R}^{H \times W \times \hat{C}}$ and $\alpha$ is a learnable temperature parameter to adjust the magnitude of inner products. \par

\textbf{Spatial Attention Layer.}
Our Spatial Attention Layer (SAL) is employed to capture spatially local and global contexts from both image and motion features. It is inspired by the spatial attention module in CBAM \cite{woo2018cbam} because their design is more lightweight as a convolution-based attention module and provides fine-grained spatial relationships between pixels. Specifically, we first use max and average pooling operations across the channel on both feature inputs. These generated 4 feature maps are concatenated and fed into a multilayer perceptron block (MLP) to generate a spatial attention map $\boldsymbol{A}$. The final output $\boldsymbol{F^{I\prime\prime}}$ after the Spatial Attention Layer is element-wise multiplication of $\boldsymbol{A}$ and input image features $\boldsymbol{F^{I\prime}}$: 
\begin{multline} \label{eq:spatialattn}
\boldsymbol{A}(\boldsymbol{F^{I\prime}}, \boldsymbol{F^m}) = \sigma (\boldsymbol{f}(MaxPool(\boldsymbol{F^{I\prime}}) \oplus AvgPool(\boldsymbol{F^{I\prime}}) \oplus \\
MaxPool(\boldsymbol{F^m}) \oplus AvgPool(\boldsymbol{F^m})))
\end{multline}
\begin{equation} \label{eq:spatialattnoutput}
\boldsymbol{F^{I\prime\prime}} = \boldsymbol{F^{I\prime}} \odot \boldsymbol{A}(\boldsymbol{F^{I\prime}}, \boldsymbol{F^m})
\end{equation}
where $\oplus$ denotes concatenation, $\sigma$ represents the activation function and $\boldsymbol{f}$ denotes a $7 \times 7$ convolution layer. \par

\begin{table*}[t]
\caption{Baseline comparisons on the low-light VSPW dataset \cite{miao2021vspw}}
\vspace{-0.6cm}
\normalsize
\center
\resizebox{1\textwidth}{!}{ 
  \begin{tabular}{cc|c|c|c|c|c|c|c}
  \toprule
  \multicolumn{2}{c|}{Method} & Backbone & mIoU $\uparrow$ & Weighted IoU $\uparrow$ & mVC\textsubscript{8} $\uparrow$ & mVC\textsubscript{16} $\uparrow$ & Params(M) $\downarrow$ & GFLOPs $\downarrow$ \\
  \hline
  Event & EV-SegNet \cite{alonso2019ev} & Xception & 18.9 & 41.2 & 78.9 & 74.4 & 29.1 & 188.6 \\
  Event & ESS \cite{sun2022ess} & E2ViD & 22.4 & 45.8 & 82.7 & 75.6 & 12.9 & 36.4 \\
  Event + Image-based & ESS \cite{sun2022ess} & E2ViD & 21.6 & 43.6 & 81.5 & 74.7 & 12.9 & 36.4 \\
  Image-based & Mask2Former \cite{cheng2022masked} & R50 & 18.1 & 42.0 & 78.3 & 73.4 & 44.0 & 110.6 \\
  Image-based & Mask2Former \cite{cheng2022masked} & Swin-T & 19.5 & 45.5 & 79.2 & 74.2 & 47.4 & 114.4 \\
  Video-based & TCB \cite{miao2021vspw} & PSPNet & 21.5 & 44.0 & 81.3 & 75.0 & 70.5 & \textemdash \\
  Video-based & TCB \cite{miao2021vspw} & OCRNet & 21.8 & 44.4 & 82.4 & 76.1 & 58.1 & \textemdash \\
  Video-based & MRCFA \cite{sun2022mining} & MiT-B0 & 16.1 & 36.1 & 74.2 & 68.9 & 5.3 & 48.2 \\
  Video-based & MRCFA \cite{sun2022mining} & MiT-B1 & 16.3 & 37.7 & 76.3 & 71.4 & 16.3 & 91.5 \\
  Video-based & CFFM \cite{sun2022coarse} & MiT-B0 & 19.9 & 46.3 & 83.0 & 75.9 & 4.7 & 26.4 \\
  Video-based & CFFM \cite{sun2022coarse} & MiT-B1 & 22.2 & 47.8 & 83.6 & 77.9 & 15.5 & 49.9 \\
  \midrule
  Video-based & \textbf{EVSNet (Ours)} & AFFormer-T & 23.6 & 52.0 & 84.9 & 79.4 & 7.4 & 30.8 \\
  Video-based & \textbf{EVSNet (Ours)} & AFFormer-B & 26.7 & 53.5 & 85.1 & 80.0 & 8.2 & 37.8 \\
  Video-based & \textbf{EVSNet (Ours)} & MiT-B0 & 28.2 & 55.7 & 87.0 & 82.1 & 9.0 & 30.1 \\
  Video-based & \textbf{EVSNet (Ours)} & MiT-B1 & 34.1 & 59.0 & 87.7 & 83.0 & 19.9 & 64.1 \\
  \bottomrule
  \end{tabular}
}
\label{table:vspw}
\end{table*}
\section{Experiments}\label{sec:experiment}
\subsection{Implementation Details}
We implement our model using MMSegmentation \cite{mmseg2020} framework and run all experiments on 2 NVIDIA RTX A5000 GPUs. For training, we use 160000 iterations with a batch size of 2. For the backbone, we adopt the Afformer (Base and Tiny) \cite{dong2023head}and MiT (B0 and B1) \cite{xie2021segformer} pre-trained on ImageNet-1K dataset \cite{deng2009imagenet}. We train the entire model using AdamW optimizer \cite{loshchilov2017decoupled} and poly learning rate schedule with the initial learning rate 6e-5. The data augmentation used in our work includes random crop, random flipping, photometric distortion, and gamma correction distortion. During training, we crop size the RGB images and event images to size $480 \times 480$ for the low-light VSPW dataset \cite{miao2021vspw}, $512 \times 1024$ for the low-light Cityscapes dataset \cite{cordts2016cityscapes} and $512 \times 512$ for the NightCity dataset \cite{tan2021night}. \par

When selecting AFFormer-Tiny and AFFormer-Base as the backbone, we set the four scales as \{1/4, 1/8, 1/8, 1/8\} of the input image spatial resolution. When using MiT-B0 and MiT-B1 as the backbone, the four scales are \{1/4, 1/8, 1/16, 1/32\} of the input image spatial resolution. \par

Follow \cite{sun2022coarse}, our model uses $\boldsymbol{l}=3$ reference frames unless otherwise specified, and $\{\boldsymbol{k_1}, \boldsymbol{k_2}, \boldsymbol{k_3}\} = \{3, 6, 9\}$. \par

\subsection{Data Preparation} 
Following previous work \cite{liang2023coherent}, we synthesize a specific low-light video frame $\boldsymbol{I_t}$ from a normal light frame $\boldsymbol{X_t}$ using linear scaling and gamma correction:
\begin{equation} \label{eq:sythesize}
\boldsymbol{I_t} = \beta \times {(\alpha \times \boldsymbol{X_t})}^\gamma
\end{equation}
where $\alpha$, $\beta$, $\gamma$ are sampled from a uniform distribution $U(0.9, 1)$, $U(0.5, 1)$, $U(2, 3.5)$, respectively. \par

To generate events based on the low-light videos, we use a popular video-to-event simulator v2e \cite{hu2021v2e}. The event images have the same spatial resolution as video frames. \par

\subsection{Datasets}
Our experiments are mainly conducted on 2 synthetic (low-light VSPW dataset \cite{miao2021vspw} and low-light Cityscapes dataset \cite{cordts2016cityscapes}) and 1 real-world datasets (NightCity dataset \cite{tan2021night}). Details of how to generate synthetic low-light dataset are described in the supplementary material. \par
\noindent \textbf{Low-light VSPW.} It has 2806 videos (198244 frames) for the training set, 343 videos (24502 frames) for the validation set, and 387 videos (28887 frames) for the test set. It selects 231 indoor and outdoor scenes and contains 124 object categories. VSPW provides the per-frame pixel-level annotations at 15 FPS which allows video scene parsing models learn the temporal information. \par
\noindent \textbf{Low-light Cityscapes.} It is a large dataset for scene understanding of urban street scenarios. Cityscapes provides pixel-level annotations per 30 frames, containing 30 object categories. Overall, it has 3475 annotated images for train/val split and 1525 annotated images for test split from collected video sequences. \par
\noindent \textbf{NightCity.} It is a large dataset with urban driving scenes at nighttime designed for supervised semantic segmentation and contains 19 categories. Overall, it consists of 4,297 real night-time images with ground truth pixel-level semantic annotations from collected video sequences. \par

\begin{table}[t]
\caption{Baseline comparisons on Low-light Cityscapes dataset \cite{cordts2016cityscapes}}
\vspace{-0.6cm}
\normalsize
\center
\resizebox{1\linewidth}{!}{ 
  \begin{tabular}{c|c|c|c|c}
  \toprule
  Method & Backbone & mIoU $\uparrow$ & Params(M) $\downarrow$ & GFLOPs $\downarrow$ \\
  \hline
  ESS \cite{sun2022ess} & E2ViD & 49.6 & 12.9 & 46.9 \\
  EV-SegNet \cite{alonso2019ev} & Xception & 41.1 & 29.1 & 245.2 \\
  MRCFA \cite{sun2022mining} & MiT-B0 & 42.1 & 5.3 & 77.5 \\
  MRCFA \cite{sun2022mining} & MiT-B1 & 45.7 & 16.3 & 145.0 \\
  CFFM \cite{sun2022coarse} & MiT-B0 & 46.0 & 4.7 & 62.4 \\
  CFFM \cite{sun2022coarse} & MiT-B1 & 50.3 & 15.5 & 118.3 \\
  \midrule
  \textbf{EVSNet (Ours)} & AFFormer-T & 57.9 & 7.4 & 70.2 \\
  \textbf{EVSNet (Ours)} & AFFormer-B & 60.9 & 8.2 & 86.0 \\
  \textbf{EVSNet (Ours)} & MiT-B0 & 59.6 & 9.0 & 70.9 \\
  \textbf{EVSNet (Ours)} & MiT-B1 & 63.2 & 19.9 & 150.1 \\
  \bottomrule
\end{tabular}}
\label{table:city}
\vspace{-0.2cm}
\end{table}

\subsection{Evaluation Metrics}
We use mean Intersection over Union (mIoU) and Weighted IoU (WIoU) to measure the per-frame segmentation performance. Weighted IoU refers to the IoU weighted by total pixel ratio of each category \cite{long2015fully}. Following \cite{miao2021vspw}, we also use Video Consistency (VC) to evaluate the temporal consistency across long-range adjacent frames category. Specifically, given a video clip with $t$ frames, ground-truth labels are $\boldsymbol{S_{1:t}}$. Assume the predicted segmentation masks are$\boldsymbol{\hat{Y}_{1:t}}$, the video consistency of is defined as:
\begin{equation} \label{eq:vc}
\vspace{-0.1cm}
VC_t = \frac{(\boldsymbol{S_1} \cap ... \cap \boldsymbol{S_t}) \cap (\boldsymbol{\hat{Y}_1} \cap ... \cap \boldsymbol{\hat{Y}_t})}{(\boldsymbol{\hat{Y}_1} \cap ... \cap \boldsymbol{\hat{Y}_t})}
\end{equation}
We use a sliding window to scan all videos with a stride of 1 and calculate the corresponding mean value $mVC_8$ and $mVC_{16}$ \par

To evaluate the model size and computational efficiency, we compare the number of parameters of the model and Giga Floating-Point Operations per Second (GFLOPS). \par

\subsection{Quantitative Results}
We evaluate the performance of our proposed EVSNet and other SOTA models on the low-light VSPW dataset in Table \ref{table:vspw}. SOTA models evaluated include Event-based models (EV-SegNet\cite{alonso2019ev}), Event+image-based models (ESS \cite{sun2022ess}), image-based models (Mask2Former\cite{cheng2022masked}), and Video-based models (TCB\cite{miao2021vspw}, CFFM\cite{sun2022coarse}, MRCFA\cite{sun2022mining}) using their default settings. 
We train these SOTA models using the training set of the low-light VSPW dataset from scratch. \par

\begin{table}[t]
\caption{Baseline comparisons on the NightCity dataset \cite{tan2021night}}
\vspace{-0.6cm}
\normalsize
\center
\resizebox{1\linewidth}{!}{ 
  \begin{tabular}{c|c|c|c}
  \toprule
  Method & Backbone & mIoU $\uparrow$ & Params(M) $\downarrow$ \\
  \hline
  DLV3P \cite{sun2022ess} & Res101 & 54.7 & 60.1 \\
  MRCFA \cite{sun2022mining} & MiT-B0 & 45.5 & 5.3 \\
  MRCFA \cite{sun2022mining} & MiT-B1 & 47.8 & 16.3 \\
  CFFM \cite{sun2022coarse} & MiT-B0 & 47.2 & 4.7 \\
  CFFM \cite{sun2022coarse} & MiT-B1 & 49.1 & 15.5 \\
  \midrule
  \textbf{EVSNet (Ours)} & MiT-B0 & 53.9 & 9.0 \\
  \textbf{EVSNet (Ours)} & MiT-B1 & 55.2 & 19.9 \\
  \bottomrule
\end{tabular}}
\label{table:night}
\vspace{-0.4cm}
\end{table}

\begin{figure*}[!t]
\centering
\includegraphics[width=1.00\textwidth]{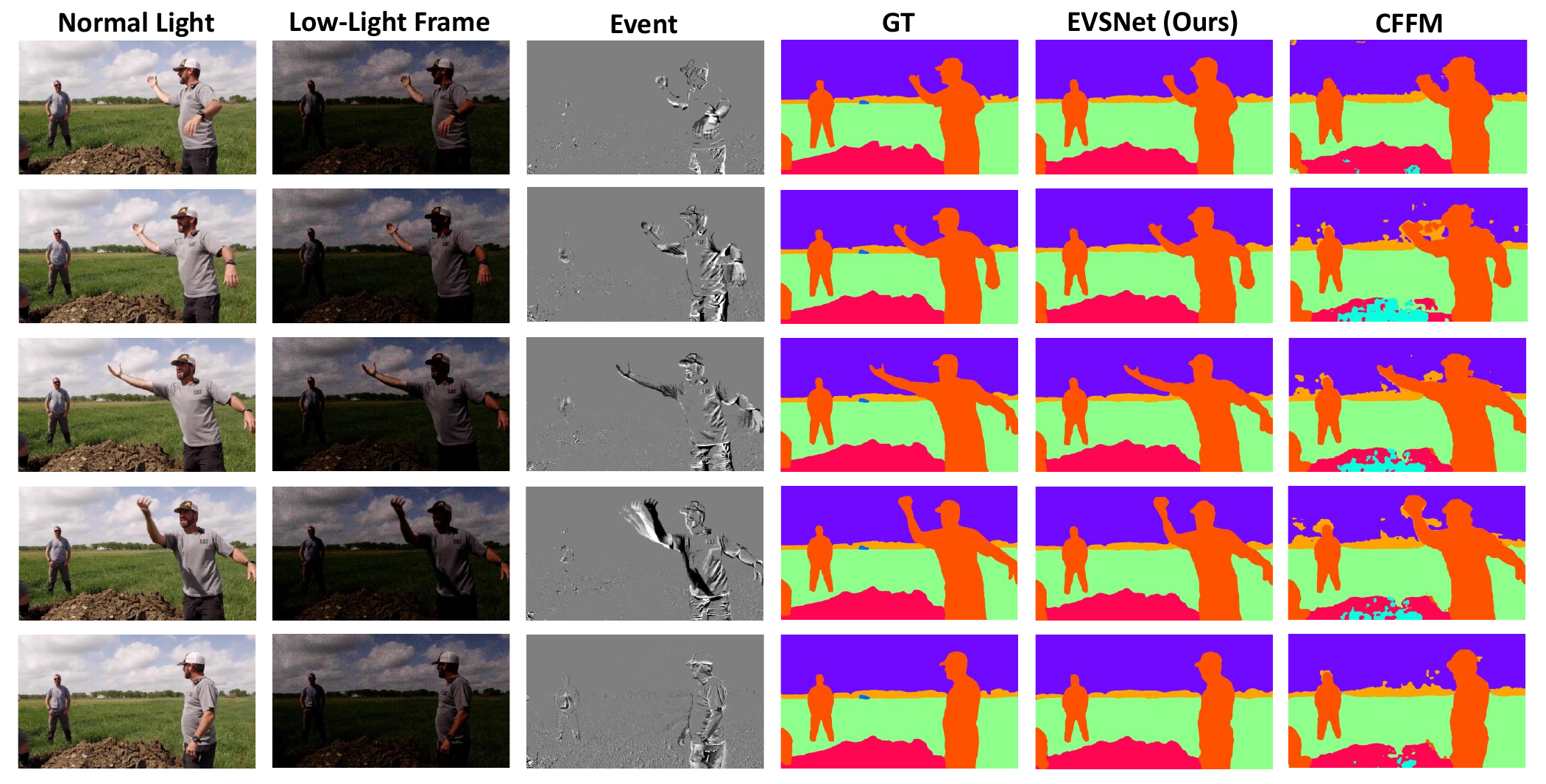}
\caption{Qualitative results on low-light VSPW dataset \cite{miao2021vspw}. From left to right: the normal light video frames, low-light video frames, the ground truth, predictions of EVSNet (ours), and predictions of CFFM \cite{sun2022coarse}. It shows that our model generates more robust and temporal consistent results, compared to the SOTA method. Best viewed in color.}
\label{vis}
\vspace{-0.4cm}
\end{figure*}

From the table, we see that EVSNet achieves an mIoU of 23.6, 26.7, 28.2, and 34.1 using the AFFormer-Tiny, AFFormer-Base, MiT-B0, and MiT-B1 backbone. It outperforms SOTA methods by a large margin on the low-light VSPW dataset. The mIoU increases by 54\% and $mVC_{16}$ increases by 7\% with similar model size. \par

We additionally compare the performance of all models using the low-light Cityscapes dataset in Table \ref{table:city}. From the table, we found that EVSNet achieves a mIoU of 57.9, 60.9, 59.6, and 63.2 using the AFFormer-Tiny, AFFormer-Base, MiT-B0, and MiT-B1 backbone, which also outperforms other SOTA models. The mIoU increases by 26\% with similar model size. \par

Similar gain is observed for NightCity in Table \ref{table:night}. EVSNet achieves a mIoU of 53.9 \& 55.2 using the MiT-B0 and MiT-B1 backbone, which shows significant improvement. The mIoU increases by 1\% with only 1/3 model size. \par

\subsection{Qualitative Results}
We visualize segmentation predictions on several video frames from the low-light VSPW dataset to better evaluate our proposed model, as shown in Figure. \ref{vis}. We compare the qualitative results of our method with one SOTA model, CFFM \cite{sun2022coarse} using its default settings. In CFFM's predictions, the category "ground" is mistakenly labeled as "stone" at the bottom, and "sky" around the person's arm is inaccurately identified as "tree", showing its struggle to recognize the object boundaries. EVSNet generates more accurate boundaries and resolves the temporal inconsistency issues of existing SOTA approaches, demonstrating the effectiveness of EVSNet. \par

\begin{table}[t]
\caption{Ablation study of the arrangement of Motion Fusion Module}
\vspace{-0.3cm}
\centering
\resizebox{1\linewidth}{!}{
  \begin{tabular}{c|c|c|c}
  \toprule
  Methods & mIOU$\uparrow$ & mVC\textsubscript{8} $\uparrow$ & mVC\textsubscript{16} $\uparrow$ \\
  \hline
   No fusion (CFFM) & 19.9 & 83.0 & 75.9 \\
   Channel & 22.5 & 84.1 & 76.3 \\
   Spatial & 21.2 & 83.7 & 75.9 \\
   Spatial + Channel & 24.5 & 86.3 & 79.4 \\
   Channel \& Spatial in parallel & 27.3 & 85.2 & 80.7\\
   Channel + Spatial (ours) & 28.2 & 87.0 & 82.1 \\
  \bottomrule
    \end{tabular}
}
\label{table:arrange}
\vspace{-0.4cm}

\end{table}

\subsection{Ablation Study}\label{sec:ablation}
\textbf{Design Choices for Motion Fusion Module}: In our proposed Motion Fusion (MFM) Module, the Channel Attention Layer and Spatial Attention Layer can be placed in parallel or sequentially. We show the ablation study results of different arrangements in Table \ref{table:arrange} using MiT-B0 backbone on the low-light VSPW dataset. Note that the no fusion option is the baseline (CFFM \cite{sun2022coarse}) results without using any event data. From our observations, both the Channel Attention Layer and Spatial Attention Layer are valuable for the proposed EVSNet. We also found that the sequential arrangement gives better result than a parallel arrangement. Further experiments also show that moving the Channel Attention Layer ahead is slightly better than having the Spatial Attention Layer first. \par
\vspace{-0.2cm}
\section{Conclusion}\label{sec:conclusion}
In this paper, we propose a novel lightweight event-guided low-light video semantic framework, EVSNet. Inspired by Atkinson-Shiffrin memory model \cite{atkinson1968human}, we leverage event modality to estimate short-term and long-term motions and further solve the video temporal inconsistency issue in low-light environments. We validate our framework using 3 large-scale datasets, low-light VSPW, low-light Cityscapes, and NightCity, and our design demonstrates significant improvements. Our results highlight the importance of effectively incorporating event features to capture motion and structural details. \par
\vspace{-0.2cm}
\section*{Acknowledgments}
This work was partially supported by a gift from Qualcomm Technologies, Inc. \par

{\small
\bibliographystyle{ieee_fullname}
\bibliography{egbib}

\begin{thebibliography}{10}\itemsep=-1pt

\bibitem{alonso2019ev}
Inigo Alonso and Ana~C Murillo.
\newblock Ev-segnet: Semantic segmentation for event-based cameras.
\newblock In {\em Proceedings of the IEEE/CVF Conference on Computer Vision and Pattern Recognition Workshops}, pages 0--0, 2019.

\bibitem{atkinson1968human}
Richard~C Atkinson and Richard~M Shiffrin.
\newblock Human memory: A proposed system and its control processes.
\newblock In {\em Psychology of learning and motivation}, volume~2, pages 89--195. Elsevier, 1968.

\bibitem{bai2024anything}
Chen Bai, Zeman Shao, Guoxiang Zhang, Di Liang, Jie Yang, Zhuorui Zhang, Yujian Guo, Chengzhang Zhong, Yiqiao Qiu, Zhendong Wang, et~al.
\newblock Anything in any scene: Photorealistic video object insertion.
\newblock {\em arXiv preprint arXiv:2401.17509}, 2024.

\bibitem{biswas2022halsie}
Shristi~Das Biswas, Adarsh Kosta, Chamika Liyanagedera, Marco Apolinario, and Kaushik Roy.
\newblock Halsie: Hybrid approach to learning segmentation by simultaneously exploiting image and event modalities.
\newblock {\em arXiv preprint arXiv:2211.10754}, 2022.

\bibitem{chen2022efficient}
Jiaan Chen, Hao Shi, Yaozu Ye, Kailun Yang, Lei Sun, and Kaiwei Wang.
\newblock Efficient human pose estimation via 3d event point cloud.
\newblock In {\em 2022 International Conference on 3D Vision (3DV)}, pages 1--10. IEEE, 2022.

\bibitem{chen2023instance}
Linwei Chen, Ying Fu, Kaixuan Wei, Dezhi Zheng, and Felix Heide.
\newblock Instance segmentation in the dark.
\newblock {\em International Journal of Computer Vision}, 131(8):2198--2218, 2023.

\bibitem{chen2019appearance}
Yadang Chen, Chuanyan Hao, Alex~X Liu, and Enhua Wu.
\newblock Appearance-consistent video object segmentation based on a multinomial event model.
\newblock {\em ACM Transactions on Multimedia Computing, Communications, and Applications (TOMM)}, 15(2):1--15, 2019.

\bibitem{cheng2022masked}
Bowen Cheng, Ishan Misra, Alexander~G Schwing, Alexander Kirillov, and Rohit Girdhar.
\newblock Masked-attention mask transformer for universal image segmentation.
\newblock In {\em Proceedings of the IEEE/CVF conference on computer vision and pattern recognition}, pages 1290--1299, 2022.

\bibitem{cheng2023tracking}
Ho~Kei Cheng, Seoung~Wug Oh, Brian Price, Alexander Schwing, and Joon-Young Lee.
\newblock Tracking anything with decoupled video segmentation.
\newblock In {\em Proceedings of the IEEE/CVF International Conference on Computer Vision}, pages 1316--1326, 2023.

\bibitem{mmseg2020}
MMSegmentation Contributors.
\newblock {MMSegmentation}: Openmmlab semantic segmentation toolbox and benchmark.
\newblock \url{https://github.com/open-mmlab/mmsegmentation}, 2020.

\bibitem{cordts2016cityscapes}
Marius Cordts, Mohamed Omran, Sebastian Ramos, Timo Rehfeld, Markus Enzweiler, Rodrigo Benenson, Uwe Franke, Stefan Roth, and Bernt Schiele.
\newblock The cityscapes dataset for semantic urban scene understanding.
\newblock In {\em Proceedings of the IEEE conference on computer vision and pattern recognition}, pages 3213--3223, 2016.

\bibitem{deng2009imagenet}
Jia Deng, Wei Dong, Richard Socher, Li-Jia Li, Kai Li, and Li Fei-Fei.
\newblock Imagenet: A large-scale hierarchical image database.
\newblock In {\em 2009 IEEE conference on computer vision and pattern recognition}, pages 248--255. Ieee, 2009.

\bibitem{dong2023head}
Bo Dong, Pichao Wang, and Fan Wang.
\newblock Head-free lightweight semantic segmentation with linear transformer.
\newblock In {\em Proceedings of the AAAI Conference on Artificial Intelligence}, volume~37, pages 516--524, 2023.

\bibitem{fan2020integrating}
Minhao Fan, Wenjing Wang, Wenhan Yang, and Jiaying Liu.
\newblock Integrating semantic segmentation and retinex model for low-light image enhancement.
\newblock In {\em Proceedings of the 28th ACM international conference on multimedia}, pages 2317--2325, 2020.

\bibitem{gadde2017semantic}
Raghudeep Gadde, Varun Jampani, and Peter~V Gehler.
\newblock Semantic video cnns through representation warping.
\newblock In {\em Proceedings of the IEEE International Conference on Computer Vision}, pages 4453--4462, 2017.

\bibitem{gehrig2023recurrent}
Mathias Gehrig and Davide Scaramuzza.
\newblock Recurrent vision transformers for object detection with event cameras.
\newblock In {\em Proceedings of the IEEE/CVF conference on computer vision and pattern recognition}, pages 13884--13893, 2023.

\bibitem{guo2024cmax}
Shuang Guo and Guillermo Gallego.
\newblock Cmax-slam: Event-based rotational-motion bundle adjustment and slam system using contrast maximization.
\newblock {\em arXiv preprint arXiv:2403.08119}, 2024.

\bibitem{hsieh2024bio}
Yung-Ting Hsieh and Dario Pompili.
\newblock A bio-inspired low-power hybrid analog/digital spiking neural networks for pervasive smart cameras.
\newblock In {\em 2024 IEEE International Conference on Pervasive Computing and Communications Workshops and other Affiliated Events (PerCom Workshops)}, pages 678--683. IEEE, 2024.

\bibitem{hu2020temporally}
Ping Hu, Fabian Caba, Oliver Wang, Zhe Lin, Stan Sclaroff, and Federico Perazzi.
\newblock Temporally distributed networks for fast video semantic segmentation.
\newblock In {\em Proceedings of the IEEE/CVF Conference on Computer Vision and Pattern Recognition}, pages 8818--8827, 2020.

\bibitem{hu2023efficient}
Yubin Hu, Yuze He, Yanghao Li, Jisheng Li, Yuxing Han, Jiangtao Wen, and Yong-Jin Liu.
\newblock Efficient semantic segmentation by altering resolutions for compressed videos.
\newblock In {\em Proceedings of the IEEE/CVF Conference on Computer Vision and Pattern Recognition}, pages 22627--22637, 2023.

\bibitem{hu2021v2e}
Yuhuang Hu, Shih-Chii Liu, and Tobi Delbruck.
\newblock v2e: From video frames to realistic dvs events.
\newblock In {\em Proceedings of the IEEE/CVF Conference on Computer Vision and Pattern Recognition}, pages 1312--1321, 2021.

\bibitem{huang2024multi}
Songjun Huang, Chuanneng Sun, Ruo-Qian Wang, and Dario Pompili.
\newblock Multi-behavior multi-agent reinforcement learning for informed search via offline training.
\newblock In {\em 2024 20th International Conference on Distributed Computing in Smart Systems and the Internet of Things (DCOSS-IoT)}, pages 19--26. IEEE, 2024.

\bibitem{huang2024toward}
Zilin Huang, Sikai Chen, Yuzhuang Pian, Zihao Sheng, Soyoung Ahn, and David~A Noyce.
\newblock Toward c-v2x enabled connected transportation system: Rsu-based cooperative localization framework for autonomous vehicles.
\newblock {\em IEEE Transactions on Intelligent Transportation Systems}, 2024.

\bibitem{jain2019accel}
Samvit Jain, Xin Wang, and Joseph~E Gonzalez.
\newblock Accel: A corrective fusion network for efficient semantic segmentation on video.
\newblock In {\em Proceedings of the IEEE/CVF Conference on Computer Vision and Pattern Recognition}, pages 8866--8875, 2019.

\bibitem{jia2022molecular}
Ruiqi Jia, Wentao Xie, Baole Wei, Guanren Qiao, Zonglin Yang, Xiaoqing Lyu, and Zhi Tang.
\newblock Molecular formula image segmentation with shape constraint loss and data augmentation.
\newblock In {\em 2022 IEEE International Conference on Bioinformatics and Biomedicine (BIBM)}, pages 3821--3823. IEEE, 2022.

\bibitem{lao2023simultaneously}
Jiangwei Lao, Weixiang Hong, Xin Guo, Yingying Zhang, Jian Wang, Jingdong Chen, and Wei Chu.
\newblock Simultaneously short-and long-term temporal modeling for semi-supervised video semantic segmentation.
\newblock In {\em Proceedings of the IEEE/CVF Conference on Computer Vision and Pattern Recognition}, pages 14763--14772, 2023.

\bibitem{lee2021gsvnet}
Shih-Po Lee, Si-Cun Chen, and Wen-Hsiao Peng.
\newblock Gsvnet: Guided spatially-varying convolution for fast semantic segmentation on video.
\newblock In {\em 2021 IEEE International Conference on Multimedia and Expo (ICME)}, pages 1--6. IEEE, 2021.

\bibitem{li2022domain}
Chenxin Li, Xin Lin, Yijin Mao, Wei Lin, Qi Qi, Xinghao Ding, Yue Huang, Dong Liang, and Yizhou Yu.
\newblock Domain generalization on medical imaging classification using episodic training with task augmentation.
\newblock {\em Computers in biology and medicine}, 141:105144, 2022.

\bibitem{li2024endora}
Chenxin Li, Hengyu Liu, Yifan Liu, Brandon~Y Feng, Wuyang Li, Xinyu Liu, Zhen Chen, Jing Shao, and Yixuan Yuan.
\newblock Endora: Video generation models as endoscopy simulators.
\newblock {\em arXiv preprint arXiv:2403.11050}, 2024.

\bibitem{li2021consistent}
Chenxin Li, Yunlong Zhang, Zhehan Liang, Wenao Ma, Yue Huang, and Xinghao Ding.
\newblock Consistent posterior distributions under vessel-mixing: a regularization for cross-domain retinal artery/vein classification.
\newblock In {\em 2021 IEEE International Conference on Image Processing (ICIP)}, pages 61--65. IEEE, 2021.

\bibitem{li2024event}
Hebei Li, Jin Wang, Jiahui Yuan, Yue Li, Wenming Weng, Yansong Peng, Yueyi Zhang, Zhiwei Xiong, and Xiaoyan Sun.
\newblock Event-assisted low-light video object segmentation.
\newblock In {\em Proceedings of the IEEE/CVF Conference on Computer Vision and Pattern Recognition}, pages 3250--3259, 2024.

\bibitem{li2021video}
Jiangtong Li, Wentao Wang, Junjie Chen, Li Niu, Jianlou Si, Chen Qian, and Liqing Zhang.
\newblock Video semantic segmentation via sparse temporal transformer.
\newblock In {\em Proceedings of the 29th ACM International Conference on Multimedia}, pages 59--68, 2021.

\bibitem{li2022video}
Xiangtai Li, Wenwei Zhang, Jiangmiao Pang, Kai Chen, Guangliang Cheng, Yunhai Tong, and Chen~Change Loy.
\newblock Video k-net: A simple, strong, and unified baseline for video segmentation.
\newblock In {\em Proceedings of the IEEE/CVF Conference on Computer Vision and Pattern Recognition}, pages 18847--18857, 2022.

\bibitem{li2021graph}
Yijin Li, Han Zhou, Bangbang Yang, Ye Zhang, Zhaopeng Cui, Hujun Bao, and Guofeng Zhang.
\newblock Graph-based asynchronous event processing for rapid object recognition.
\newblock In {\em Proceedings of the IEEE/CVF International Conference on Computer Vision}, pages 934--943, 2021.

\bibitem{li2024feature}
Zhenglin Li, Yangchen Huang, Mengran Zhu, Jingyu Zhang, JingHao Chang, and Houze Liu.
\newblock Feature manipulation for ddpm based change detection.
\newblock {\em arXiv preprint arXiv:2403.15943}, 2024.

\bibitem{liang2023coherent}
Jinxiu Liang, Yixin Yang, Boyu Li, Peiqi Duan, Yong Xu, and Boxin Shi.
\newblock Coherent event guided low-light video enhancement.
\newblock In {\em Proceedings of the IEEE/CVF International Conference on Computer Vision}, pages 10615--10625, 2023.

\bibitem{liao2024mobile}
Youqi Liao, Shuhao Kang, Jianping Li, Yang Liu, Yun Liu, Zhen Dong, Bisheng Yang, and Xieyuanli Chen.
\newblock Mobile-seed: Joint semantic segmentation and boundary detection for mobile robots.
\newblock {\em IEEE Robotics and Automation Letters}, 2024.

\bibitem{long2015fully}
Jonathan Long, Evan Shelhamer, and Trevor Darrell.
\newblock Fully convolutional networks for semantic segmentation.
\newblock In {\em Proceedings of the IEEE conference on computer vision and pattern recognition}, pages 3431--3440, 2015.

\bibitem{loshchilov2017decoupled}
Ilya Loshchilov and Frank Hutter.
\newblock Decoupled weight decay regularization.
\newblock {\em arXiv preprint arXiv:1711.05101}, 2017.

\bibitem{miao2021vspw}
Jiaxu Miao, Yunchao Wei, Yu Wu, Chen Liang, Guangrui Li, and Yi Yang.
\newblock Vspw: A large-scale dataset for video scene parsing in the wild.
\newblock In {\em Proceedings of the IEEE/CVF conference on computer vision and pattern recognition}, pages 4133--4143, 2021.

\bibitem{nguyen2017real}
Anh Nguyen, Thanh-Toan Do, Darwin~G Caldwell, and Nikos~G Tsagarakis.
\newblock Real-time pose estimation for event cameras with stacked spatial lstm networks.
\newblock {\em arXiv preprint arXiv:1708.09011}, 3, 2017.

\bibitem{nilsson2018semantic}
David Nilsson and Cristian Sminchisescu.
\newblock Semantic video segmentation by gated recurrent flow propagation.
\newblock In {\em Proceedings of the IEEE conference on computer vision and pattern recognition}, pages 6819--6828, 2018.

\bibitem{paul2021local}
Matthieu Paul, Martin Danelljan, Luc Van~Gool, and Radu Timofte.
\newblock Local memory attention for fast video semantic segmentation.
\newblock In {\em 2021 IEEE/RSJ International Conference on Intelligent Robots and Systems (IROS)}, pages 1102--1109. IEEE, 2021.

\bibitem{paul2020efficient}
Matthieu Paul, Christoph Mayer, Luc~Van Gool, and Radu Timofte.
\newblock Efficient video semantic segmentation with labels propagation and refinement.
\newblock In {\em Proceedings of the IEEE/CVF Winter Conference on Applications of Computer Vision}, pages 2873--2882, 2020.

\bibitem{qiao2024multi}
Guanren Qiao, Guiliang Liu, Pascal Poupart, and Zhiqiang Xu.
\newblock Multi-modal inverse constrained reinforcement learning from a mixture of demonstrations.
\newblock {\em Advances in Neural Information Processing Systems}, 36, 2024.

\bibitem{qiu2023sats}
Yiqiao Qiu, Yixing Shen, Zhuohao Sun, Yanchong Zheng, Xiaobin Chang, Weishi Zheng, and Ruixuan Wang.
\newblock Sats: Self-attention transfer for continual semantic segmentation.
\newblock {\em Pattern Recognition}, 138:109383, 2023.

\bibitem{sun2023hmaac}
Chuanneng Sun, Songjun Huang, and Dario Pompili.
\newblock Hmaac: Hierarchical multi-agent actor-critic for aerial search with explicit coordination modeling.
\newblock In {\em 2023 IEEE International Conference on Robotics and Automation (ICRA)}, pages 7728--7734. IEEE, 2023.

\bibitem{sun2022coarse}
Guolei Sun, Yun Liu, Henghui Ding, Thomas Probst, and Luc Van~Gool.
\newblock Coarse-to-fine feature mining for video semantic segmentation.
\newblock In {\em proceedings of the IEEE/CVF conference on computer vision and pattern recognition}, pages 3126--3137, 2022.

\bibitem{sun2022mining}
Guolei Sun, Yun Liu, Hao Tang, Ajad Chhatkuli, Le Zhang, and Luc Van~Gool.
\newblock Mining relations among cross-frame affinities for video semantic segmentation.
\newblock In {\em European Conference on Computer Vision}, pages 522--539. Springer, 2022.

\bibitem{sun2022ess}
Zhaoning Sun, Nico Messikommer, Daniel Gehrig, and Davide Scaramuzza.
\newblock Ess: Learning event-based semantic segmentation from still images.
\newblock In {\em European Conference on Computer Vision}, pages 341--357. Springer, 2022.

\bibitem{tan2021night}
Xin Tan, Ke Xu, Ying Cao, Yiheng Zhang, Lizhuang Ma, and Rynson~WH Lau.
\newblock Night-time scene parsing with a large real dataset.
\newblock {\em IEEE Transactions on Image Processing}, 30:9085--9098, 2021.

\bibitem{vitale2021event}
Antonio Vitale, Alpha Renner, Celine Nauer, Davide Scaramuzza, and Yulia Sandamirskaya.
\newblock Event-driven vision and control for uavs on a neuromorphic chip.
\newblock In {\em 2021 IEEE International Conference on Robotics and Automation (ICRA)}, pages 103--109. IEEE, 2021.

\bibitem{wang2021temporal}
Hao Wang, Weining Wang, and Jing Liu.
\newblock Temporal memory attention for video semantic segmentation.
\newblock In {\em 2021 IEEE International Conference on Image Processing (ICIP)}, pages 2254--2258. IEEE, 2021.

\bibitem{wang2021evdistill}
Lin Wang, Yujeong Chae, Sung-Hoon Yoon, Tae-Kyun Kim, and Kuk-Jin Yoon.
\newblock Evdistill: Asynchronous events to end-task learning via bidirectional reconstruction-guided cross-modal knowledge distillation.
\newblock In {\em Proceedings of the IEEE/CVF Conference on Computer Vision and Pattern Recognition}, pages 608--619, 2021.

\bibitem{wang2024unsupervised}
Wenjing Wang, Rundong Luo, Wenhan Yang, and Jiaying Liu.
\newblock Unsupervised illumination adaptation for low-light vision.
\newblock {\em IEEE Transactions on Pattern Analysis \& Machine Intelligence}, (01):1--15, 2024.

\bibitem{wang2023dual}
Ziyang Wang and Congying Ma.
\newblock Dual-contrastive dual-consistency dual-transformer: A semi-supervised approach to medical image segmentation.
\newblock In {\em Proceedings of the IEEE/CVF International Conference on Computer Vision}, pages 870--879, 2023.

\bibitem{wang2022ev}
Ziyun Wang, Fernando~Cladera Ojeda, Anthony Bisulco, Daewon Lee, Camillo~J Taylor, Kostas Daniilidis, M~Ani Hsieh, Daniel~D Lee, and Volkan Isler.
\newblock Ev-catcher: High-speed object catching using low-latency event-based neural networks.
\newblock {\em IEEE Robotics and Automation Letters}, 7(4):8737--8744, 2022.

\bibitem{wang2024mixsegnet}
Ziyang Wang and Chen Yang.
\newblock Mixsegnet: Fusing multiple mixed-supervisory signals with multiple views of networks for mixed-supervised medical image segmentation.
\newblock {\em Engineering Applications of Artificial Intelligence}, 133:108059, 2024.

\bibitem{wang2022metateacher}
Zhenbin Wang, Mao Ye, Xiatian Zhu, Liuhan Peng, Liang Tian, and Yingying Zhu.
\newblock Metateacher: Coordinating multi-model domain adaptation for medical image classification.
\newblock {\em Advances in Neural Information Processing Systems}, 35:20823--20837, 2022.

\bibitem{wang2024landa}
Zhenbin Wang, Lei Zhang, Lituan Wang, and Minjuan Zhu.
\newblock Landa: Language-guided multi-source domain adaptation.
\newblock {\em arXiv preprint arXiv:2401.14148}, 2024.

\bibitem{wei2022lidar}
Yi Wei, Zibu Wei, Yongming Rao, Jiaxin Li, Jie Zhou, and Jiwen Lu.
\newblock Lidar distillation: Bridging the beam-induced domain gap for 3d object detection.
\newblock In {\em European Conference on Computer Vision}, pages 179--195. Springer, 2022.

\bibitem{woo2018cbam}
Sanghyun Woo, Jongchan Park, Joon-Young Lee, and In~So Kweon.
\newblock Cbam: Convolutional block attention module.
\newblock In {\em Proceedings of the European conference on computer vision (ECCV)}, pages 3--19, 2018.

\bibitem{wu2021one}
Xinyi Wu, Zhenyao Wu, Lili Ju, and Song Wang.
\newblock A one-stage domain adaptation network with image alignment for unsupervised nighttime semantic segmentation.
\newblock {\em IEEE Transactions on Pattern Analysis and Machine Intelligence}, 45(1):58--72, 2021.

\bibitem{xia2023cmda}
Ruihao Xia, Chaoqiang Zhao, Meng Zheng, Ziyan Wu, Qiyu Sun, and Yang Tang.
\newblock Cmda: Cross-modality domain adaptation for nighttime semantic segmentation.
\newblock In {\em Proceedings of the IEEE/CVF International Conference on Computer Vision}, pages 21572--21581, 2023.

\bibitem{xie2021segformer}
Enze Xie, Wenhai Wang, Zhiding Yu, Anima Anandkumar, Jose~M Alvarez, and Ping Luo.
\newblock Segformer: Simple and efficient design for semantic segmentation with transformers.
\newblock {\em Advances in neural information processing systems}, 34:12077--12090, 2021.

\bibitem{xu2018dynamic}
Yu-Syuan Xu, Tsu-Jui Fu, Hsuan-Kung Yang, and Chun-Yi Lee.
\newblock Dynamic video segmentation network.
\newblock In {\em Proceedings of the IEEE conference on computer vision and pattern recognition}, pages 6556--6565, 2018.

\bibitem{yang2021focal}
Jianwei Yang, Chunyuan Li, Pengchuan Zhang, Xiyang Dai, Bin Xiao, Lu Yuan, and Jianfeng Gao.
\newblock Focal self-attention for local-global interactions in vision transformers.
\newblock {\em arXiv preprint arXiv:2107.00641}, 2021.

\bibitem{yang2024monogae}
Lei Yang, Xinyu Zhang, Jiaxin Yu, Jun Li, Tong Zhao, Li Wang, Yi Huang, Chuang Zhang, Hong Wang, and Yiming Li.
\newblock Monogae: Roadside monocular 3d object detection with ground-aware embeddings.
\newblock {\em IEEE Transactions on Intelligent Transportation Systems}, 2024.

\bibitem{yao2024cracknex}
Zhen Yao, Jiawei Xu, Shuhang Hou, and Mooi~Choo Chuah.
\newblock Cracknex: a few-shot low-light crack segmentation model based on retinex theory for uav inspections.
\newblock {\em arXiv preprint arXiv:2403.03063}, 2024.

\bibitem{ying2021srnet}
Xiaowen Ying, Xin Li, and Mooi~Choo Chuah.
\newblock Srnet: Spatial relation network for efficient single-stage instance segmentation in videos.
\newblock In {\em Proceedings of the 29th ACM International Conference on Multimedia}, pages 347--356, 2021.

\bibitem{zamir2022restormer}
Syed~Waqas Zamir, Aditya Arora, Salman Khan, Munawar Hayat, Fahad~Shahbaz Khan, and Ming-Hsuan Yang.
\newblock Restormer: Efficient transformer for high-resolution image restoration.
\newblock In {\em Proceedings of the IEEE/CVF conference on computer vision and pattern recognition}, pages 5728--5739, 2022.

\bibitem{zhang2022lisu}
Ning Zhang, Francesco Nex, Norman Kerle, and George Vosselman.
\newblock Lisu: Low-light indoor scene understanding with joint learning of reflectance restoration.
\newblock {\em ISPRS journal of photogrammetry and remote sensing}, 183:470--481, 2022.

\bibitem{zhao2017pyramid}
Hengshuang Zhao, Jianping Shi, Xiaojuan Qi, Xiaogang Wang, and Jiaya Jia.
\newblock Pyramid scene parsing network.
\newblock In {\em Proceedings of the IEEE conference on computer vision and pattern recognition}, pages 2881--2890, 2017.

\bibitem{zhao2024road}
Tong Zhao, Yichen Xie, Mingyu Ding, Lei Yang, Masayoshi Tomizuka, and Yintao Wei.
\newblock A road surface reconstruction dataset for autonomous driving.
\newblock {\em Scientific data}, 11(1):459, 2024.

\bibitem{zhao2024roadbev}
Tong Zhao, Lei Yang, Yichen Xie, Mingyu Ding, Masayoshi Tomizuka, and Yintao Wei.
\newblock Roadbev: Road surface reconstruction in bird's eye view.
\newblock {\em arXiv preprint arXiv:2404.06605}, 2024.

\bibitem{zhou2018semi}
Yi Zhou, Guillermo Gallego, Henri Rebecq, Laurent Kneip, Hongdong Li, and Davide Scaramuzza.
\newblock Semi-dense 3d reconstruction with a stereo event camera.
\newblock In {\em Proceedings of the European conference on computer vision (ECCV)}, pages 235--251, 2018.

\bibitem{zhu2023topic}
Yicheng Zhu, Yiqiao Qiu, Qingyuan Wu, Fu~Lee Wang, and Yanghui Rao.
\newblock Topic driven adaptive network for cross-domain sentiment classification.
\newblock {\em Information Processing \& Management}, 60(2):103230, 2023.

\end{thebibliography}
}

\end{document}